\pdfoutput=1

\documentclass[11pt]{article}

\usepackage[]{acl}
\usepackage{times}
\usepackage{latexsym}

\usepackage[T1]{fontenc}

\usepackage[utf8]{inputenc}

\usepackage{microtype}

\usepackage{inconsolata}

\usepackage{enumerate}
\usepackage{amsmath,amssymb}
\usepackage{cleveref}
\usepackage{booktabs}
\usepackage{graphicx}
\usepackage{multicol,multirow}
\usepackage[boldmath]{numprint}
\usepackage{subfig}
\usepackage{comment}
\usepackage[export]{adjustbox}
\usepackage{relsize}
\usepackage{ifthen}
\usepackage{tikz}
\usetikzlibrary{positioning}
\usetikzlibrary{calc}
\usetikzlibrary{arrows.meta}
\usetikzlibrary{shapes}
\usetikzlibrary{decorations.pathreplacing}
\usepackage{pbox}

\usepackage{booktabs,colortbl,xcolor}
\usepackage{siunitx}  
\usepackage[normalem]{ulem}

\usepackage{ragged2e}
\title{I Have an Attention Bridge to Sell You: \\ Generalization Capabilities of Modular Translation Architectures}

\author{
    Timothee Mickus \\ \And Ra\'ul V\'azquez \\ \And Joseph Attieh \\[-1.0cm] 
\AND \null \\[-1.0cm] University of Helsinki \\ { \tt firstname.lastname@helsinki.fi}
}

\begin{document}
\maketitle
\begin{abstract}
Modularity is a paradigm of machine translation with the potential of bringing forth models that are large at training time and small during inference.
Within this field of study, modular approaches, and in particular attention bridges, have been argued to improve the generalization capabilities of models by fostering language-independent representations. 
In the present paper, we study whether modularity affects translation quality; as well as how well modular architectures generalize across different evaluation scenarios.
For a given computational budget, we find non-modular architectures to be always comparable or preferable to all modular designs we study.
\end{abstract}

\section{Introduction}
Machine Translation (MT) has historically been under two influences that seem \textsl{a prima facie} contradictory.
One of the goals of MT research is to provide means of converting sentences from any language to any other.
On the one hand, generalization capabilities hinge on our systems producing language agnostic representations.
On the other hand, MT models ought to be apt at encoding the specifics of source languages \citep{belinkov-etal-2017-neural}.
The former of these trends has deeply marked this field---the concept of an `interlingua' runs through most of the history of MT research, from \citet{richens-56-preprogramming} to \citet{lu-etal-2018-neural}.
The latter has recently motivated the development of modular approaches, where network parameters are specifically tied to a specific language.

How can we reconcile these two seemingly paradoxical trends?
One promising approach is the inclusion of fully-shared subnetworks in modular architectures, and especially \emph{bridge} components: 
They have been argued to foster language-independent representations \citep{zhu-etal-2020-language} as well as zero-shot generalization capabilities \citep{liao-etal-2021-improving}.
Our aim is to carefully assess whether modular architectures in general and bridges do indeed foster greater generalization capabilities.

We therefore study six architectures, five of which modular, with a particular focus on how they generalize---both to unseen translation directions, and to novel domains.
We find that 
modular systems still struggle to remain competitive with fully-shared MT systems in scenarios when not all translation directions are available---a conclusion that affects systems with and without fixed-size bridges equally.
While encoder-sharing modular designs can rival or outperform non-modular settings in a wide range of scenarios, all other systems we study struggle in zero-shot and out-of-distribution conditions, strongly questioning that fully-shared sub-networks in modular MT systems can improve their generalization capabilities.


\section{Related Work}


The full span of multilingual NMT (MNMT) architectures rely in the implicit assumption that the systems leverage the multilingual data by creating a shared encoding space via sharing: from fully-shared models 
\citep{johnson-etal-2017-googles}, to fully-modular systems, where sharing occurs only at dataset level \citep{escolano-etal-2021-multilingual}. In this work, we assess those two extreme cases, focusing in the modular NMT systems that incorporate some parameter-sharing bridging layers. 
\citet{lu-etal-2018-neural} introduced  an attentional neural interlingua, which processes language-specific encoder embeddings to produce language-agnostic representations. 
\citet{zhu-etal-2020-language} proposed a language-aware interlingua that transforms the encoder representation to a shared semantic space, showcasing practical means of fostering the semantic consistency of translations.  \citet{vazquez-etal-2019-multilingual} integrated a shared inner-attention mechanism, referred to as ``attention bridge'', based on the work of \citet{lin2017a}, to generate fixed-size sentence representations. Further studies by \citet{raganato-etal-2019-evaluation} and \citet{vazquez-etal-2020-systematic}, whose work we specifically build upon, emphasized the advantages of using multiple attention heads on the semantic quality of the translation---as well as challenges,  particularly with translating longer sentences. \citet{boggia-etal-2023-dozens} explored the effects of sharing encoder parameters vs. increasing the number of languages in modular MNMT.  
More recently, \citet{purason-tattar-2022-multilingual} used layers shared by language groups to enhance translation,  \citet{mao2023variablelength} proposed a variable-length bridge that uses a classification layer to predict its length, and in \citet{pires2023learning} the encoder is built with interspersed fully-shared and language-specific layers.

\section{Experimental Methodology}
\label{sec:methodology}

\subsection{Model Variants}
\label{sec:methodology:archs}

\begin{figure*}[t]
    \centering
    \input{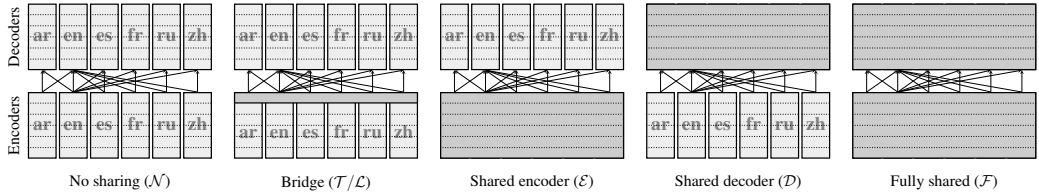}
    \caption{Overview of considered architectures, focusing on \textbf{EN} setting (using English as a pivot). Layers shaded in dark gray are shared across all languages; layers shaded in light gray are specific to a source or target language.}
    \label{fig:archs}
\end{figure*}


All the models we consider are Transformer-based  \citep{vaswani-etal-2017-attention}, and implemented with the MAMMOTH library \citep{mickus-etal-2024-mammoth}.\footnote{
    Configuration files available at \href{https://github.com/Helsinki-NLP/mammoth/tree/main/examples/ab-neg}{\tt github.com/
    Helsinki-NLP/mammoth/tree/main/examples/ab-neg/}.
}
An overview of the different modular architectures we consider is displayed in \Cref{fig:archs}.
We ensure that all datapoints are processed by the same number of encoder and decoder layers (6 and 6 resp.).

\paragraph{Non-modular baseline.} To provide a reasonable point of comparison with existing approaches, we consider a simple non-modular architecture where all parameters are shared across all translation directions.
We note these fully-shared models as $\mathcal{F}$.

\paragraph{Fully modular baseline.} A second natural point of comparison is a modular system without bridge; e.g. \citet{escolano-etal-2021-multilingual}.
Such models, noted $\mathcal{N}$ below, contain one 6-layer Transformer encoder and one 6-layer Transformer decoder per language, which are then selected for predictions depending on the desired language pair.

\paragraph{Semi-modular approaches.} All other remaining architectures we will discuss contain both language specific and language-independent parameters.
A simple means of achieving this consist in using a single shared encoder for all source languages (abbrv. $\mathcal{E}$), which would allow to leverage training signals from all source languages so as to provide more robust encoder representations.
Conversely, one can consider employing a single shared decoder for al target languages (abbrv. $\mathcal{D}$) in the hopes of bolstering generation capabilities.

\paragraph{Bridges.} We also consider models with a ``bridge'' layer, i.e., where all parameters are language specific aside from the last Transformer layer in the encoder.
Such models have been explored by e.g. \citet{boggia-etal-2023-dozens}. 
These models are noted $\mathcal{T}$, and contain 5-layer language-specific Transformer encoders, followed by a shared Transformer layer serving as a  bridge---i.e. they  are $\mathcal{N}$-type modular systems where the parameters of the last layers of each encoder are tied.

\paragraph{Fixed-size attention bridges.}
An alternative proposed by \citet{vazquez-etal-2020-systematic} consists in using fixed-size attention bridge (FSAB) designs.
FSAB models, noted $\mathcal{L}$, resemble $\mathcal{T}$ models except for the fact that the fully-shared Transformer layer bridge is replaced by the structured embedding architecture proposed by \citet{lin2017a}: 
\begin{equation} \label{eq:lin}
    \mathbf{Y} = \operatorname{softmax}\left(\mathbf{W}_Q^{{}} \operatorname{ReLU} \left(\mathbf{W}_K^{{}} \mathbf{X}\right)^{\top}\right) \cdot \mathbf{X}
\end{equation}
with 
$\mathbf{X}$ 
the input matrix 
of the shared layer.
Models of the $\mathcal{L}$ architecture contain language-specific encoders comprising 5 Transformer layers, followed by one FSAB layer shared across all languages.

\subsection{Datasets}
\label{sec:methodology:datasets}

We use two MT datasets:
the United Nations Parallel Corpus \citep[UNPC]{ziemski-etal-2016-united}, which contains documents in six UN languages (Arabic, Chinese, English, French, Russian, and Spanish); and OPUS100 \citep{zhang-etal-2020-improving}, an English-centric 
multilingual corpus derived from \citet{tiedemann-2012-parallel} spanning 100 languages.
%
We ignore all OPUS translation directions not present in UNPC.
Since the UNPC contains over 10M paired sentences across six languages (Arabic, English, Spanish, French, Russian, Mandarin Chinese), we consider the entire released data, rather than the fully aligned sub-corpus, and hold out 10\% of the data for any evaluation and/or experiments. 
We ensure that sentences are unique to a split, i.e., if a pair of sentences $\left( s_1, s_2 \right)$ is present in the test split, then any pair $\left( s_1, s_3\right)$ involving either of these sentence will also be assigned to the test split.
Out of these 10\%, we randomly select 25k sentences per language pairs to use as test sets. 
The remaining 90\% examples are used for training, with 10k sentences per language pairs set aside for validation. 

\paragraph{Test splits for generalization.}
We assess generalization capabilities in two common setups: zero-shot translation directions and out-of-distribution (OOD) examples. 
To evaluate out-of-distribution performances, we simply train models on one dataset (UNPC or OPUS) and evaluate it on the other (resp. OPUS or UNPC). 
Since bridge components are argued to be useful for unseen translation directions \citep{liao-etal-2021-improving}, we experiment with different language pivots to artificially create zero-shot translation directions. 
We construct three distinct UNPC training sets: (i) one using all 30 translation directions available in the UNPC, (``\textbf{All}''); (ii) one using all 10 directions involving English as a source or target (``\textbf{EN}''); and (iii) one using all 10 directions involving Arabic as a source or target (``\textbf{AR}''). 
This allows us to evaluate our models in both English-centric and non-English-centric contexts as well as in a zero-shot setting.\footnote{
    Since OPUS100 is English-centric, only one variant of this dataset is considered for training.
} Hence, we refer to \textbf{EN} or \textbf{AR} being pivot languages, when an experiment is centered around that language.


\paragraph{Training conditions.}
To enable zero-shot translation \citep[cf.]{vazquez-etal-2019-multilingual,artetxe-schwenk-2019-massively}, we train our models on auto-encoding tasks for all 6 languages.
UNPC models are trained on monolingual data derived from the UNPC, and likewise OPUS models are trained on OPUS monolingual data.
We train three seeds of all six model variants ($\mathcal{F}$, $\mathcal{N}$, $\mathcal{E}$, $\mathcal{D}$, $\mathcal{T}$, $\mathcal{L}$) on the four training sets (\textbf{UNPC-All}, \textbf{UNPC-EN} \textbf{UNPC-AR}, \textbf{OPUS-EN}) under a strictly controlled computational budget: 
All models are exposed to the same number of datapoints and are trained with 6 AMD MI250X GPUs. 
We use the hyperparameters of \citet{boggia-etal-2023-dozens} aside from batch accumulation, set to 8. 
We use $k=50$ in $\mathcal{L}$ models as \citet{vazquez-etal-2020-systematic}.

\begin{table*}[t!]
    \centering

   \begin{adjustbox}{max width=\linewidth, max height=0.1825\textheight}
            \sisetup{
separate-uncertainty,
table-align-uncertainty,
table-format = 2.1(1),
detect-weight=true}
\begin{tabular}{p{0.5cm}@{{\,\,}}p{0.75cm}@{{\,\,}}l@{{\,}}l  >{\columncolor{gray!25}}S@{{\quad}}S@{{\quad}}>{\columncolor{gray!25}}S@{{\quad}}S@{{\quad}}>{\columncolor{gray!25}}S@{{\quad}}S }
&& \multicolumn{2}{@{{}}c}{\multirow{2}{*}{\pbox{1.75cm}{\textbf{Translation} \newline \textbf{directions}}}} &&&&& \\ 
&&&& $\mathcal{N}$ & $\mathcal{F}$ & $\mathcal{E}$ & $\mathcal{D}$ & $\mathcal{T}$ & $\mathcal{L}$  \\
\toprule
\multirow{12}{*}{\rotatebox{90}{test on \textbf{UNPC}}}
&\multirow{7}{*}{\rotatebox{90}{train on \textbf{UNPC}}}
& \textbf{All} & (seen) & 26.6(0.5) & \bf 28.2(1.1) &  \uline{$\mathbf{29.0 \pm 0.2}$} & 24.8(0.2) & 26.6(0.1) & 26.4(0.2) \\
\cmidrule(lr){3-10}
&&\multirow{3}{*}{\textbf{AR}} 
& all & 18.1(0.3) &  \uline{$\mathbf{24.6 \pm 1.3}$} & \bf 23.1(1.7) & 15.7(1.8) & 16.8(0.4) & 17.9(0.1) \\
&&& seen & 26.4(0.1) & \uline{$\mathbf{27.1\pm 0.9}$} & \bf 26.6(0.7) & 22.0(1.0) & 24.9(0.1) & {$26.2 \pm 0.0$} \\
&&& unseen & 13.9(0.3) &  \uline{$\mathbf{23.4\pm 1.4}$} & \bf 21.4(2.3) & 12.5(2.2) & 12.8(0.6) & 13.7(0.1) \\
\cmidrule(lr){3-10}
&&\multirow{3}{*}{\textbf{EN}} 
& all & 19.3(0.4) & \bf 22.8(2.9) &  \uline{$\mathbf{23.9\pm 1.0}$} & 17.9(0.4) & 19.3(0.1) & 19.4(0.1) \\
&&& seen & 34.5(0.2) & 33.1(2.6) &  \uline{$\mathbf{35.9\pm 0.1}$} & 31.6(0.9) & 34.0(0.2) & \bf 34.6(0.3) \\
&&& unseen & 11.7(00.6) &  \bf 17.6(3.0) &  \uline{$\mathbf{17.9\pm 1.4}$} & 11.0(1.0) & 11.9(0.1) & 11.8(0.1) \\
\cmidrule{2-10}
&\multirow{3}{*}{\rotatebox{90}{\pbox{1.75cm}{train on \\ \textbf{OPUS}}}}
&\multirow{3}{*}{\textbf{EN}}
& all & 16.4(0.2) & \bf 20.6(0.5) &  \uline{$\mathbf{20.9 \pm 0.4}$} & 13.6(1.2) & 16.8(0.2) & 16.3(0.1) \\
&&& seen  & \bf 30.8(0.2) & 30.5(0.5) &  \uline{$\mathbf{31.1 \pm 0.3}$} & 23.7(1.5) & 30.7(0.3) & 30.6(0.3) \\
&&& unseen & 9.1(0.3) & \bf 15.6(0.5) &  \uline{$\mathbf{15.8 \pm 0.5}$} & 8.5(1.0) & 9.9(0.1) & 9.1(0.1) \\

\midrule
\multirow{12}{*}{\rotatebox{90}{test on \textbf{OPUS}}}
&\multirow{7}{*}{\rotatebox{90}{train on \textbf{UNPC}}}
& \textbf{All} & (seen) & 17.6(0.2) & \bf 19.1(0.8) & \uline{$\mathbf{19.7 \pm 0.2}$} & 16.3(0.2) & 17.5(0.2) & 17.5(0.3) \\
\cmidrule(lr){3-10}
&&\multirow{3}{*}{\textbf{AR}} 
& all & 12.3(0.2) & \uline{$\mathbf{16.5 \pm 0.8}$} & \bf 15.4(1.0) & 10.3(1.4) & 11.4(0.2) & 12.1(0.1) \\
&&& seen & 17.7(0.1) & \uline{$\mathbf{18.4 \pm 0.8}$} & \bf 17.9(0.6) & 13.8(1.0) & 16.6(0.1) & 17.6(0.1)  \\
&&& unseen  & 9.2(0.3) & \uline{$\mathbf{15.5 \pm 0.8}$} & \bf 13.9(1.3) & 8.4(1.6) & 8.5(0.2) & 9.0(0.1) \\
\cmidrule(lr){3-10}
&&\multirow{3}{*}{\textbf{EN}} 
& all & 13.4(0.3) & \bf 15.9(2.0) & \uline{$\mathbf{17.0 \pm 0.5}$} & 12.6(0.2) & 13.3(0.1) & 13.5(0.2) \\
&&& seen  & 19.7(0.1) & \bf 19.8(1.4) & \uline{$\mathbf{21.0 \pm 0.0}$} & 18.5(0.7) & 19.2(0.1) & \bf 19.8(0.2) \\
&&& unseen & 8.2(0.3) & \bf 12.7(2.5) & \uline{$\mathbf{13.6 \pm 0.9}$} & 7.7(0.7) & 8.4(0.3) & 8.2(0.3)  \\
\cmidrule{2-10}
&\multirow{3}{*}{\rotatebox{90}{\pbox{1.75cm}{train on \\ \textbf{OPUS}}}}
&\multirow{3}{*}{\textbf{EN}}
& all  & 15.0(0.2) & \bf 17.8(0.3) & \uline{$\mathbf{17.9 \pm 0.3}$} & 12.4(1.0) & 15.5(0.2) & 14.8(0.1) \\
&&& seen  & \uline{$\mathbf{25.1 \pm 0.2}$} & 24.6(0.3) & \uline{$\mathbf{25.1 \pm 0.2}$} & 19.7(1.6) & 24.9(0.2) & 24.9(0.3)  \\
&&& unseen & 6.6(0.2) & \uline{$\mathbf{12.1 \pm 0.3}$} & \bf 11.8(0.5) & 6.3(0.7) & 7.7(0.2) & 6.4(0.1) \\
\bottomrule
\end{tabular}

    \end{adjustbox}
    \caption{Summary of performances, with \underline{\textbf{best}} and \textbf{second best} values highlighted (avg. of 3 seeds $\pm$ std. dev.), and broken down according to whether the translation direction was seen  during training or not (i.e., zero shot).}
    \label{tab:table1}
\end{table*}

\section{Results}
\label{sec:results}

The primary metric used for evaluating the performance of our models is BLEU \citep{papineni-etal-2002-bleu,post-2018-call}.\footnote{
    While COMET \citep{rei-etal-2020-comet} would in principle be preferable, computing it for all translation directions in every model in our study is prohibitively costly. 
}
Results are shown in \Cref{tab:table1}. 
 


\paragraph{Choice of architecture.} 
A clear trend emerges from our results: 
Across the board, the encoder-shared models $\mathcal{E}$ are found to be the most successful, followed by the fully-shared, non-modular models $\mathcal{F}$. 
The latter only prevails upon the former in Arabic-centric scenario.
At times, these architectures outrank other models considered by large margins of up to 7.5 BLEU points. 
While fully modular $\mathcal{N}$ models or FSAB-based $\mathcal{L}$ models perform well in the \textbf{EN}-centric scenario, these are not overwhelmingly better than $\mathcal{F}$.

\paragraph{Choice of pivot language.}
We experiment with different pivot languages, \textbf{EN} and \textbf{AR}, to understand their influence on the results. 
Our observations indicate that the choice of a pivot language can significantly impact the outcomes: 
The results with \textbf{AR} are always below the corresponding scores with \textbf{EN} on translation directions studied during training, whereas \textbf{AR} models yield generally higher performance in zero-shot conditions than their \textbf{EN} counterparts. 
Furthermore, we find tentative evidence that the behavior in \textbf{EN} and \textbf{AR} differs from that of \textbf{All}:
In the latter case, we find a more limited impact of the architecture being used, with score varying at most by $\pm 4.2$ BLEU points; whereas we observe a spread of up to $\pm 7.3$ BLEU points for the former.  
As one would expect, being exposed to all translation directions during training (\textbf{All}) allows to improve performances averaged all translation directions.
If we restrict ourselves to directions a model was exposed to during training, we find that \textbf{EN} models often outperform \textbf{All} models; whereas \textbf{AR} models are more in line with the values we see for \textbf{All}. 
This would suggest that there is a difficulty inherent to the translation directions considered; focusing only on directions that involve English may inflate performances.

\paragraph{Translation directions (seen vs. unseen).}
Expanding on what we already briefly touched on, we systematically find performances in zero-shot conditions to remain firmly below what we observe for translation directions observed during training.
This holds across pivot languages and architectures.
We do not observe that bridges ($\mathcal{T}$ or $\mathcal{L}$) provide benefits in terms of zero-shot performances over fully modular systems ($\mathcal{N}$). 
instead, it would appear that sharing the encoder ($\mathcal{E}$ or $\mathcal{F}$) is beneficial---although it is uncertain that this is due to greater generalization capabilities rather than overall improved performances, the improvement brought about by $\mathcal{E}$ and $\mathcal{F}$ models is more substantiated in zero-shot settings (with a gap of at least $4.1$ BLEU points in zero-shot settings, whereas $\mathcal{F}$ can be outperformed by $\mathcal{L}$ and/or $\mathcal{N}$ for training directions).

\paragraph{In-distribution vs. out-of-distribution.}
Comparing performances in-distribution and out-of-distribution does not suggest that bridges meaningfully improve generalization capabilities.
Performances of $\mathcal{T}$ and $\mathcal{L}$ models are  in line with what we observe for the bridge-less $\mathcal{N}$ models.

\section{Statistical modeling} 
\paragraph{SHAP analysis \& predictors importance}
%
Are our observations statistically significant?
To establish which factors are at play, we 
rely on SHAP \citep{NIPS2017_8a20a862}, a library and algorithm to derive heuristics for Shapley values \citep{shapley:book1952}.
We fit a gradient boosting decision tree regression model with CatBoost \citep{NEURIPS2018_14491b75} to explain the BLEU scores obtained on specific language pairs and datasets by all the models we trained.
We use as predictors (i) the source language (categorical); (ii) the target language (categorical), (iii) whether the model was trained on UNPC (binary); (iv) whether this translation direction in zero-shot (binary); (v) whether this test corresponds to an out-of-distribution setting (binary);  as well as (vi--xi) which architecture is used (binary predicates for each of $\mathcal{N}$, $\mathcal{F}$, $\mathcal{E}$, $\mathcal{D}$, $\mathcal{T}$, and $\mathcal{L}$).

\begin{figure}[t]
    \centering
    
        \includegraphics[max width=0.95\linewidth]{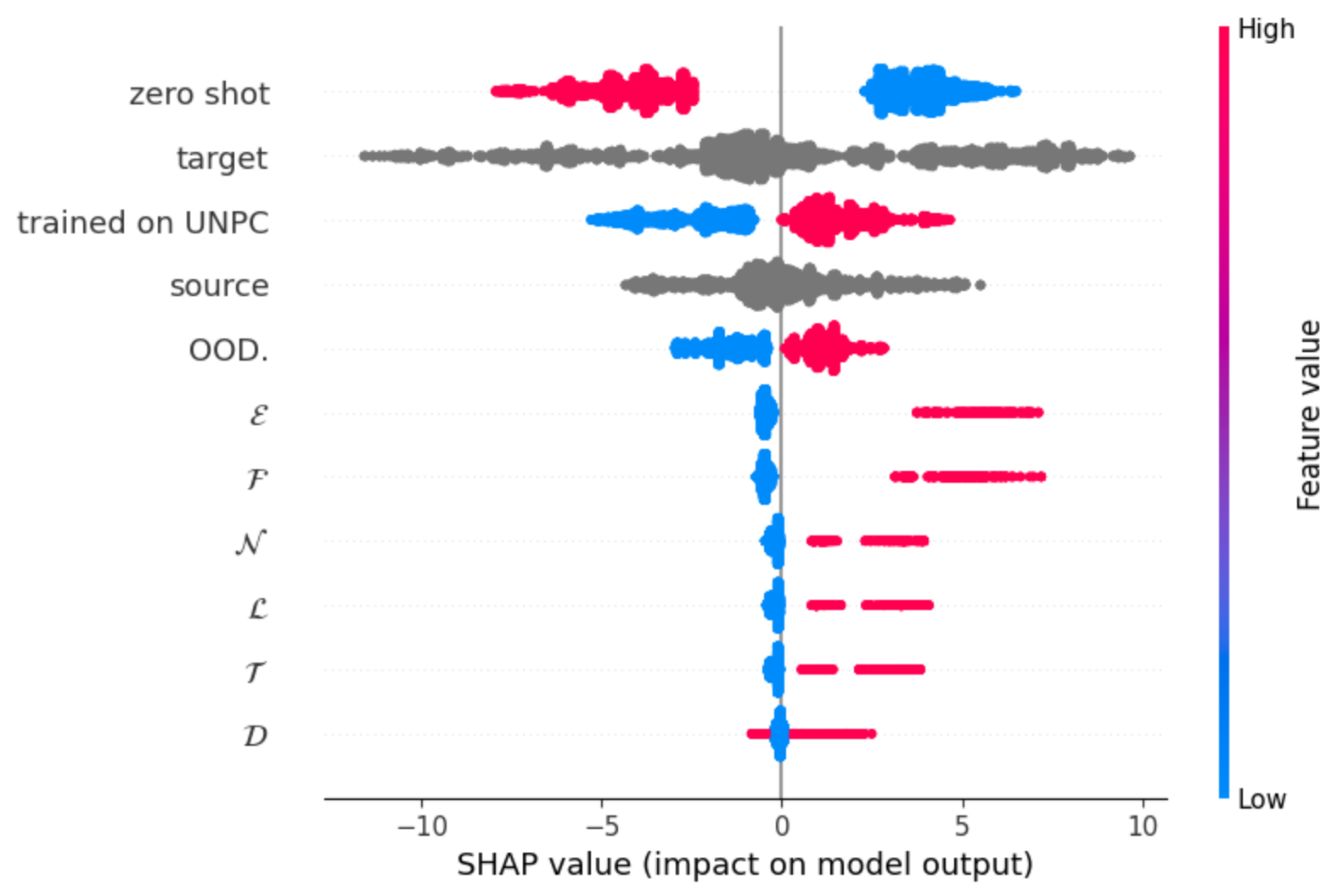}
    \caption{Overview of SHAP values, sorted by mean absolute value. Grey: categorical predictors; red: binary predictors where the value is true; blue, where it is false.}
    \label{fig:shap}
\end{figure}

\Cref{fig:shap} provides a general overview of the results of this analysis. 
The exact evaluation conditions---i.e. the training and testing corpora and the specific language pairs seen at training and during the test at hand all, corresponding to predictors (i--v)---have a strong impact on the observed BLEU scores.
We also see that using models of type $\mathcal{F}$ and $\mathcal{E}$ more strongly and more positively impacts the BLEU scores we observe than any other model type.
In short, 
we find that most modular models fail to bring about results  comparable with what we see for our non-modular baseline $\mathcal{F}$, with the sole exception of encoder-sharing $\mathcal{E}$.

\begin{table}[t]
\sisetup{
separate-uncertainty,
table-format = 3.4,
detect-weight=true}
    \centering
\resizebox{\linewidth}{!}{
\begin{tabular}{lSSSS}
\toprule
                              & \textbf{coef} & \textbf{std err} & \textbf{t} & \textbf{P$> |$t$|$} \\
\midrule
\textit{Intercept}            &      11.8312  &        0.211     &    56.153  &         0.000       \\
\textbf{has bridge}           &       4.4287  &        0.248     &    17.851  &         0.000       \\
\textbf{shares enc}           &       5.3919  &        0.248     &    21.733  &         0.000       \\
\textbf{zero shot}            &      -7.5567  &        0.139     &   -54.277  &         0.000       \\
\textbf{OOD}                  &       1.9472  &        0.140     &    13.917  &         0.000       \\
\textbf{from EN}              &       1.9792  &        0.178     &    11.099  &         0.000       \\
\textbf{from ES}              &       2.7001  &        0.204     &    13.235  &         0.000       \\
\textbf{from FR}              &       1.5625  &        0.182     &     8.578  &         0.000       \\
\textbf{from RU}              &       0.5932  &        0.182     &     3.257  &         0.001       \\
\textbf{from ZH}              &      -2.6625  &        0.182     &   -14.617  &         0.000       \\
\textbf{to EN}                &       9.1120  &        0.178     &    51.098  &         0.000       \\
\textbf{to ES}                &       7.7651  &        0.204     &    38.061  &         0.000       \\
\textbf{to FR}                &       4.0147  &        0.182     &    22.040  &         0.000       \\
\textbf{to RU}                &       2.0937  &        0.182     &    11.494  &         0.000       \\
\textbf{to ZH}                &      -5.2907  &        0.182     &   -29.046  &         0.000       \\
\textbf{trained on UNPC}      &       4.3278  &        0.125     &    34.705  &         0.000       \\
\midrule
\textbf{has bridge$\times$zero shot} &      -4.7733  &        0.283     &   -16.876  &         0.000       \\
\textbf{has bridge$\times$OOD}       &       0.4169  &        0.283     &     1.472  &         0.141       \\
\textbf{shares enc$\times$zero shot} &       0.3607  &        0.283     &     1.275  &         0.202       \\
\textbf{shares enc$\times$OOD}       &       0.9785  &        0.283     &     3.455  &         0.001       \\
\bottomrule
\end{tabular}
}
    \caption{OLS coefficients and significance. Intercept: $\mathcal{N}$-type, not OOD, not zero-shot, from \& to AR.}
    \label{tab:ols}
\end{table}

\paragraph{OLS model \& predictors interaction.}
Is there evidence that some modular architectures (and bridges in particular) enhance generalization capabilities?
While SHAP values 
provide 
independent coefficients for each 
factor,
this question is at its core one of interrelation
---and is thus best studied through models able to capture potential interactions between predictors.
To that end, we fit a simple ordinary least squares (OLS) linear model to predict the BLEU scores of our models 
using as predictors (i) whether the architecture contains a bridge (i.e., models of type $\mathcal{T}$ or $\mathcal{L}$); (ii) whether it shares the encoder across source languages  (i.e., models of type $\mathcal{F}$ or $\mathcal{E}$); (iii) whether the model is tested in zero-shot; (iv) whether it is tested in an OOD setting; (v) whether the model was trained on UNPC; (vi \& vii) the source and target languages; (viii--xi) the interactions between modular design (i.e., predictors i \& ii) and performances in generalization conditions (viz. predictors iii \& iv).\footnote{
   We ignore datapoints from type $\mathcal{D}$ models since we are not aware of specific claims with respect to this architecture. 
}

Our model achieves a $R^2$ of $0.763$. Predictor coefficients and significance are listed in \Cref{tab:ols}.
As expected, modular design and training \& test conditions (predictors i--vii) are always significant.
Zero shot performances are linked to the strongest negative coefficient in our model; likewise, translating from or to ZH also turns out to degrade performance somewhat compared to the intercept (AR).
Looking at interactions,
we find that models with a bridge require a clear \emph{negative} correction in zero-shot scenarios, \emph{opposite} to what has been argued by \citet{liao-etal-2021-improving}.
M
odels of type $\mathcal{F}$ and $\mathcal{E}$ require a positive correction in OOD settings, suggesting they distinguish themselves further from other modular architectures.
%
This statistical modeling suggests that bridge-based architectures significantly decrease generalization capabilities, as opposed to other modular ($\mathcal{E}$) and non-modular ($\mathcal{F}$) designs---in contrast with much of the discourse about their benefit for language independence and usefulness in zero-shot conditions \citep{raganato-etal-2019-evaluation,zhu-etal-2020-language,vazquez-etal-2020-systematic}.

\section{Conclusions}
\label{sec:ccl}

In this work, we study the claim that bridge layers in modular architectures foster greater generalization capabilities.
Given a carefully controlled computational budget, bridge architectures never clearly outperform 
bridge-less architectures, be they modular or not.
In particular, we find non-modular architectures exhibit strong competitiveness, as they are only outperformed by modular architectures with language independent encoders and modular language-specific decoders.
Additionally, we note that training conditions, such as the translation direction accessible to a model during training, have a significant impact.

These results suggest that current modular architectures, especially those using bridging layers, have limited potential insofar MT is concerned. 
In most cases, a default non-modular transformer fares better or just as well than the most effective modular system.
Our 
study 
focused on
modular architectures in a small-scale, well controlled experimental protocol; we leave questions such as whether these remarks carry on at a larger scale, both of model parameter counts and number of languages concerned, for future work.

\section*{Acknowledgements}
\vspace{1ex}
\noindent
{ 
\begin{minipage}{0.1\linewidth}
    \vspace{-10pt}
    \raisebox{-0.2\height}{\includegraphics[trim =32mm 55mm 30mm 5mm, clip, scale=0.18]{figs/logos/erc.ai}} \\[0.25cm]
    \raisebox{-0.25\height}{\includegraphics[trim =0mm 5mm 5mm 2mm,clip,scale=0.075]{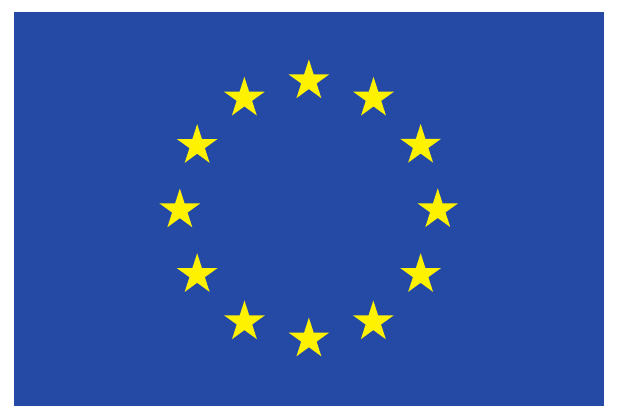}}
\end{minipage}
\hspace{0.01\linewidth}
\begin{minipage}{0.85\linewidth}
This work is part of the FoTran project, funded by the European Research Council (ERC) under the EU's Horizon 2020 research and innovation program (agreement \textnumero{}~771113). \hfill We ~also ~thank ~the ~CSC-IT\vspace{0.5ex}
\end{minipage}
\begin{minipage}{\linewidth}
\noindent Center for Science Ltd., for computational resources and NVIDIA AI Technology Center (NVAITC) for the expertise in distributed training.
\end{minipage}%
}

\bibliography{custom,anthology}

\begin{thebibliography}{27}
\expandafter\ifx\csname natexlab\endcsname\relax\def\natexlab#1{#1}\fi

\bibitem[{Artetxe and Schwenk(2019)}]{artetxe-schwenk-2019-massively}
Mikel Artetxe and Holger Schwenk. 2019.
\newblock \href {https://doi.org/10.1162/tacl_a_00288} {Massively multilingual
  sentence embeddings for zero-shot cross-lingual transfer and beyond}.
\newblock \emph{Transactions of the Association for Computational Linguistics},
  7:597--610.

\bibitem[{Belinkov et~al.(2017)Belinkov, Durrani, Dalvi, Sajjad, and
  Glass}]{belinkov-etal-2017-neural}
Yonatan Belinkov, Nadir Durrani, Fahim Dalvi, Hassan Sajjad, and James Glass.
  2017.
\newblock \href {https://doi.org/10.18653/v1/P17-1080} {What do neural machine
  translation models learn about morphology?}
\newblock In \emph{Proceedings of the 55th Annual Meeting of the Association
  for Computational Linguistics (Volume 1: Long Papers)}, pages 861--872,
  Vancouver, Canada. Association for Computational Linguistics.

\bibitem[{Boggia et~al.(2023)Boggia, Gr{\"o}nroos, Loppi, Mickus, Raganato,
  Tiedemann, and V{\'a}zquez}]{boggia-etal-2023-dozens}
Michele Boggia, Stig-Arne Gr{\"o}nroos, Niki Loppi, Timothee Mickus, Alessandro
  Raganato, J{\"o}rg Tiedemann, and Ra{\'u}l V{\'a}zquez. 2023.
\newblock \href {https://aclanthology.org/2023.nodalida-1.24} {Dozens of
  translation directions or millions of shared parameters? comparing two types
  of multilinguality in modular machine translation}.
\newblock In \emph{Proceedings of the 24th Nordic Conference on Computational
  Linguistics (NoDaLiDa)}, pages 238--247, T{\'o}rshavn, Faroe Islands.
  University of Tartu Library.

\bibitem[{Escolano et~al.(2021)Escolano, Costa-juss{\`a}, Fonollosa, and
  Artetxe}]{escolano-etal-2021-multilingual}
Carlos Escolano, Marta~R. Costa-juss{\`a}, Jos{\'e} A.~R. Fonollosa, and Mikel
  Artetxe. 2021.
\newblock \href {https://doi.org/10.18653/v1/2021.eacl-main.80} {Multilingual
  machine translation: Closing the gap between shared and language-specific
  encoder-decoders}.
\newblock In \emph{Proceedings of the 16th Conference of the European Chapter
  of the Association for Computational Linguistics: Main Volume}, pages
  944--948, Online. Association for Computational Linguistics.

\bibitem[{Johnson et~al.(2017)Johnson, Schuster, Le, Krikun, Wu, Chen, Thorat,
  Vi{\'e}gas, Wattenberg, Corrado, Hughes, and
  Dean}]{johnson-etal-2017-googles}
Melvin Johnson, Mike Schuster, Quoc~V. Le, Maxim Krikun, Yonghui Wu, Zhifeng
  Chen, Nikhil Thorat, Fernanda Vi{\'e}gas, Martin Wattenberg, Greg Corrado,
  Macduff Hughes, and Jeffrey Dean. 2017.
\newblock \href {https://doi.org/10.1162/tacl_a_00065} {{G}oogle{'}s
  multilingual neural machine translation system: Enabling zero-shot
  translation}.
\newblock \emph{Transactions of the Association for Computational Linguistics},
  5:339--351.

\bibitem[{Liao et~al.(2021)Liao, Shi, Gong, Shou, Qu, and
  Zeng}]{liao-etal-2021-improving}
Junwei Liao, Yu~Shi, Ming Gong, Linjun Shou, Hong Qu, and Michael Zeng. 2021.
\newblock \href {https://doi.org/10.1109/IJCNN52387.2021.9534401} {Improving
  zero-shot neural machine translation on language-specific encoders-
  decoders}.
\newblock In \emph{2021 International Joint Conference on Neural Networks
  (IJCNN)}, pages 1--8.

\bibitem[{Lin et~al.(2017)Lin, Feng, dos Santos, Yu, Xiang, Zhou, and
  Bengio}]{lin2017a}
Zhouhan Lin, Minwei Feng, Cicero~Nogueira dos Santos, Mo~Yu, Bing Xiang, Bowen
  Zhou, and Yoshua Bengio. 2017.
\newblock \href {https://openreview.net/forum?id=BJC_jUqxe} {A structured
  self-attentive sentence embedding}.
\newblock In \emph{International Conference on Learning Representations}.

\bibitem[{Lu et~al.(2018)Lu, Keung, Ladhak, Bhardwaj, Zhang, and
  Sun}]{lu-etal-2018-neural}
Yichao Lu, Phillip Keung, Faisal Ladhak, Vikas Bhardwaj, Shaonan Zhang, and
  Jason Sun. 2018.
\newblock \href {https://doi.org/10.18653/v1/W18-6309} {A neural interlingua
  for multilingual machine translation}.
\newblock In \emph{Proceedings of the Third Conference on Machine Translation:
  Research Papers}, pages 84--92, Brussels, Belgium. Association for
  Computational Linguistics.

\bibitem[{Lundberg and Lee(2017)}]{NIPS2017_8a20a862}
Scott~M Lundberg and Su-In Lee. 2017.
\newblock \href
  {https://proceedings.neurips.cc/paper_files/paper/2017/file/8a20a8621978632d76c43dfd28b67767-Paper.pdf}
  {A unified approach to interpreting model predictions}.
\newblock In \emph{Advances in Neural Information Processing Systems},
  volume~30. Curran Associates, Inc.

\bibitem[{Mao et~al.(2023)Mao, Song, Dabre, Chu, and
  Kurohashi}]{mao2023variablelength}
Zhuoyuan Mao, Haiyue Song, Raj Dabre, Chenhui Chu, and Sadao Kurohashi. 2023.
\newblock \href {http://arxiv.org/abs/2305.10190} {Variable-length neural
  interlingua representations for zero-shot neural machine translation}.

\bibitem[{Mickus et~al.(2024)Mickus, Gr{\"o}nroos, Attieh, Boggia, De~Gibert,
  Ji, Loppi, Raganato, V{\'a}zquez, and Tiedemann}]{mickus-etal-2024-mammoth}
Timothee Mickus, Stig-Arne Gr{\"o}nroos, Joseph Attieh, Michele Boggia, Ona
  De~Gibert, Shaoxiong Ji, Niki~Andreas Loppi, Alessandro Raganato, Ra{\'u}l
  V{\'a}zquez, and J{\"o}rg Tiedemann. 2024.
\newblock \href {https://aclanthology.org/2024.eacl-demo.14} {{MAMMOTH}:
  Massively multilingual modular open translation @ {H}elsinki}.
\newblock In \emph{Proceedings of the 18th Conference of the European Chapter
  of the Association for Computational Linguistics: System Demonstrations},
  pages 127--136, St. Julians, Malta. Association for Computational
  Linguistics.

\bibitem[{Papineni et~al.(2002)Papineni, Roukos, Ward, and
  Zhu}]{papineni-etal-2002-bleu}
Kishore Papineni, Salim Roukos, Todd Ward, and Wei-Jing Zhu. 2002.
\newblock \href {https://doi.org/10.3115/1073083.1073135} {{B}leu: a method for
  automatic evaluation of machine translation}.
\newblock In \emph{Proceedings of the 40th Annual Meeting of the Association
  for Computational Linguistics}, pages 311--318, Philadelphia, Pennsylvania,
  USA. Association for Computational Linguistics.

\bibitem[{Pires et~al.(2023)Pires, Schmidt, Liao, and
  Peitz}]{pires2023learning}
Telmo~Pessoa Pires, Robin~M. Schmidt, Yi-Hsiu Liao, and Stephan Peitz. 2023.
\newblock \href {http://arxiv.org/abs/2305.02665} {Learning language-specific
  layers for multilingual machine translation}.

\bibitem[{Post(2018)}]{post-2018-call}
Matt Post. 2018.
\newblock \href {https://doi.org/10.18653/v1/W18-6319} {A call for clarity in
  reporting {BLEU} scores}.
\newblock In \emph{Proceedings of the Third Conference on Machine Translation:
  Research Papers}, pages 186--191, Brussels, Belgium. Association for
  Computational Linguistics.

\bibitem[{Prokhorenkova et~al.(2018)Prokhorenkova, Gusev, Vorobev, Dorogush,
  and Gulin}]{NEURIPS2018_14491b75}
Liudmila Prokhorenkova, Gleb Gusev, Aleksandr Vorobev, Anna~Veronika Dorogush,
  and Andrey Gulin. 2018.
\newblock \href
  {https://proceedings.neurips.cc/paper_files/paper/2018/file/14491b756b3a51daac41c24863285549-Paper.pdf}
  {Catboost: unbiased boosting with categorical features}.
\newblock In \emph{Advances in Neural Information Processing Systems},
  volume~31. Curran Associates, Inc.

\bibitem[{Purason and T{\"a}ttar(2022)}]{purason-tattar-2022-multilingual}
Taido Purason and Andre T{\"a}ttar. 2022.
\newblock \href {https://aclanthology.org/2022.eamt-1.12} {Multilingual neural
  machine translation with the right amount of sharing}.
\newblock In \emph{Proceedings of the 23rd Annual Conference of the European
  Association for Machine Translation}, pages 91--100, Ghent, Belgium. European
  Association for Machine Translation.

\bibitem[{Raganato et~al.(2019)Raganato, V{\'a}zquez, Creutz, and
  Tiedemann}]{raganato-etal-2019-evaluation}
Alessandro Raganato, Ra{\'u}l V{\'a}zquez, Mathias Creutz, and J{\"o}rg
  Tiedemann. 2019.
\newblock \href {https://doi.org/10.18653/v1/W19-4304} {An evaluation of
  language-agnostic inner-attention-based representations in machine
  translation}.
\newblock In \emph{Proceedings of the 4th Workshop on Representation Learning
  for NLP (RepL4NLP-2019)}, pages 27--32, Florence, Italy. Association for
  Computational Linguistics.

\bibitem[{Rei et~al.(2020)Rei, Stewart, Farinha, and
  Lavie}]{rei-etal-2020-comet}
Ricardo Rei, Craig Stewart, Ana~C Farinha, and Alon Lavie. 2020.
\newblock \href {https://doi.org/10.18653/v1/2020.emnlp-main.213} {{COMET}: A
  neural framework for {MT} evaluation}.
\newblock In \emph{Proceedings of the 2020 Conference on Empirical Methods in
  Natural Language Processing (EMNLP)}, pages 2685--2702, Online. Association
  for Computational Linguistics.

\bibitem[{Richens(1956)}]{richens-56-preprogramming}
Richard~H. Richens. 1956.
\newblock \href
  {https://aclanthology.org/www.mt-archive.info/50/MT-1956-Richens.pdf}
  {Preprogramming for mechanical translation}.
\newblock \emph{Mechanical Translation}, 3(1):20--25.

\bibitem[{Shapley(1953)}]{shapley:book1952}
Lloyd~S Shapley. 1953.
\newblock A value for n-person games.
\newblock In Harold~W. Kuhn and Albert~W. Tucker, editors, \emph{Contributions
  to the Theory of Games II}, pages 307--317. Princeton University Press,
  Princeton.

\bibitem[{Tiedemann(2012)}]{tiedemann-2012-parallel}
J{\"o}rg Tiedemann. 2012.
\newblock \href
  {http://www.lrec-conf.org/proceedings/lrec2012/pdf/463_Paper.pdf} {Parallel
  data, tools and interfaces in {OPUS}}.
\newblock In \emph{Proceedings of the Eighth International Conference on
  Language Resources and Evaluation ({LREC}'12)}, pages 2214--2218, Istanbul,
  Turkey. European Language Resources Association (ELRA).

\bibitem[{Vaswani et~al.(2017)Vaswani, Shazeer, Parmar, Uszkoreit, Jones,
  Gomez, Kaiser, and Polosukhin}]{vaswani-etal-2017-attention}
Ashish Vaswani, Noam Shazeer, Niki Parmar, Jakob Uszkoreit, Llion Jones,
  Aidan~N Gomez, \L~ukasz Kaiser, and Illia Polosukhin. 2017.
\newblock \href
  {https://proceedings.neurips.cc/paper/2017/file/3f5ee243547dee91fbd053c1c4a845aa-Paper.pdf}
  {Attention is all you need}.
\newblock In \emph{Advances in Neural Information Processing Systems},
  volume~30. Curran Associates, Inc.

\bibitem[{V{\'a}zquez et~al.(2020)V{\'a}zquez, Raganato, Creutz, and
  Tiedemann}]{vazquez-etal-2020-systematic}
Ra{\'u}l V{\'a}zquez, Alessandro Raganato, Mathias Creutz, and J{\"o}rg
  Tiedemann. 2020.
\newblock \href {https://doi.org/10.1162/coli_a_00377} {A systematic study of
  inner-attention-based sentence representations in multilingual neural machine
  translation}.
\newblock \emph{Computational Linguistics}, 46(2):387--424.

\bibitem[{V{\'a}zquez et~al.(2019)V{\'a}zquez, Raganato, Tiedemann, and
  Creutz}]{vazquez-etal-2019-multilingual}
Ra{\'u}l V{\'a}zquez, Alessandro Raganato, J{\"o}rg Tiedemann, and Mathias
  Creutz. 2019.
\newblock \href {https://doi.org/10.18653/v1/W19-4305} {Multilingual {NMT} with
  a language-independent attention bridge}.
\newblock In \emph{Proceedings of the 4th Workshop on Representation Learning
  for NLP (RepL4NLP-2019)}, pages 33--39, Florence, Italy. Association for
  Computational Linguistics.

\bibitem[{Zhang et~al.(2020)Zhang, Williams, Titov, and
  Sennrich}]{zhang-etal-2020-improving}
Biao Zhang, Philip Williams, Ivan Titov, and Rico Sennrich. 2020.
\newblock \href {https://doi.org/10.18653/v1/2020.acl-main.148} {Improving
  massively multilingual neural machine translation and zero-shot translation}.
\newblock In \emph{Proceedings of the 58th Annual Meeting of the Association
  for Computational Linguistics}, pages 1628--1639, Online. Association for
  Computational Linguistics.

\bibitem[{Zhu et~al.(2020)Zhu, Yu, Cheng, and Luo}]{zhu-etal-2020-language}
Changfeng Zhu, Heng Yu, Shanbo Cheng, and Weihua Luo. 2020.
\newblock \href {https://doi.org/10.18653/v1/2020.acl-main.150} {Language-aware
  interlingua for multilingual neural machine translation}.
\newblock In \emph{Proceedings of the 58th Annual Meeting of the Association
  for Computational Linguistics}, pages 1650--1655, Online. Association for
  Computational Linguistics.

\bibitem[{Ziemski et~al.(2016)Ziemski, Junczys-Dowmunt, and
  Pouliquen}]{ziemski-etal-2016-united}
Micha{\l} Ziemski, Marcin Junczys-Dowmunt, and Bruno Pouliquen. 2016.
\newblock \href {https://aclanthology.org/L16-1561} {The {U}nited {N}ations
  parallel corpus v1.0}.
\newblock In \emph{Proceedings of the Tenth International Conference on
  Language Resources and Evaluation ({LREC}'16)}, pages 3530--3534,
  Portoro{\v{z}}, Slovenia. European Language Resources Association (ELRA).

\end{thebibliography}

\end{document}